\documentclass{article}
\usepackage{spconf,amsmath,graphicx}
\usepackage{color}
\usepackage{amsfonts,amsthm}
\usepackage{multirow}
\usepackage{lipsum}
\usepackage{paralist}
\usepackage{wasysym}
\usepackage{mathrsfs}
\usepackage{multirow}
\usepackage{microtype}
\usepackage{booktabs}

\title{Modeling Intention, Emotion and External World in Dialogue Systems}

\name{Wei Peng \textsuperscript{1,2}, Yue Hu\textsuperscript{1,2}\sthanks{Corresponding author. E-mail: huyue@iie.ac.cn}, Luxi Xing\textsuperscript{1,2}, Yuqiang Xie\textsuperscript{1,2}, Xingsheng Zhang\textsuperscript{1,2}, Yajing Sun\textsuperscript{1,2}}
\address{\textsuperscript{1}Institute of Information Engineering, Chinese Academy of Sciences, China \\
\textsuperscript{2}School of Cyber Security, University of Chinese Academy of Sciences, China}
\begin{document}
%\ninept
%

\maketitle

\begin{abstract}
Intention, emotion and action are important elements in human activities. Modeling the interaction process between individuals by analyzing the relationships between these elements is a challenging task.
% In this paper, we propose a novel Cognitive Interaction Framework (CogInf) for intention recognition and emotion prediction tasks. To verify the effectiveness of the framework, we propose two corresponding models to recognize intention and emotion. 
However, previous work mainly focused on modeling intention and emotion independently, and neglected of exploring the mutual relationships between intention and emotion. In this paper, we propose a RelAtion Interaction Network (RAIN), consisting of Intention Relation Module and Emotion Relation Module, to jointly model mutual relationships and explicitly integrate historical intention information. 
The experiments on the dataset show that our model can take full advantage of the intention, emotion and action between individuals and achieve a remarkable improvement over BERT-style baselines. Qualitative analysis verifies the importance of the mutual interaction between the intention and emotion. % The novel framework shows an interesting perspective on human interaction in cognitive science.
\end{abstract}
\section{Introduction}
\label{sec:intro}
%\blfootnote{
%	Copyright 2021 IEEE. Published in ICASSP 2021 - 2021 IEEE International Conference on Acoustics, Speech and Signal Processing (ICASSP), scheduled for 6-11 June 2021 in Toronto, Ontario, Canada. Personal use of this material is permitted. However, permission to reprint/republish this material for advertising or promotional purposes or for creating new collective works for resale or redistribution to servers or lists, or to reuse any copyrighted component of this work in other works, must be obtained from the IEEE. Contact: Manager, Copyrights and Permissions / IEEE Service Center / 445 Hoes Lane / P.O. Box 1331 / Piscataway, NJ 08855-1331, USA. Telephone: + Intl. 908-562-3966.}
%Services, Inc.: Phone +1-979-846-6800 or email
%to \\\texttt{papers@2021.ieeeicassp.org}.
Intention recognition and emotion prediction are long-term researches in dialogue systems \cite{1997Affective, maslow2013theory}. Intention \cite{Gable2010Approach, reeve2014understanding}, as an essential psychological background to stimulate and guide action, is the internal dynamic tendency to carry out activities, thus affecting the state of the external world. For example, when a person wants to purchase, he will have a conversation with the shop assistant to ask to know the information about the goods, including material, price, size, etc. At this time, the intention of the customer can be also inferred to purchase from the act of dialogue. On the contrary, he will not communicate with the shop assistant if the customer has no intention.

\begin{figure}[t]
	\centering
	\includegraphics[width=0.38\textwidth]{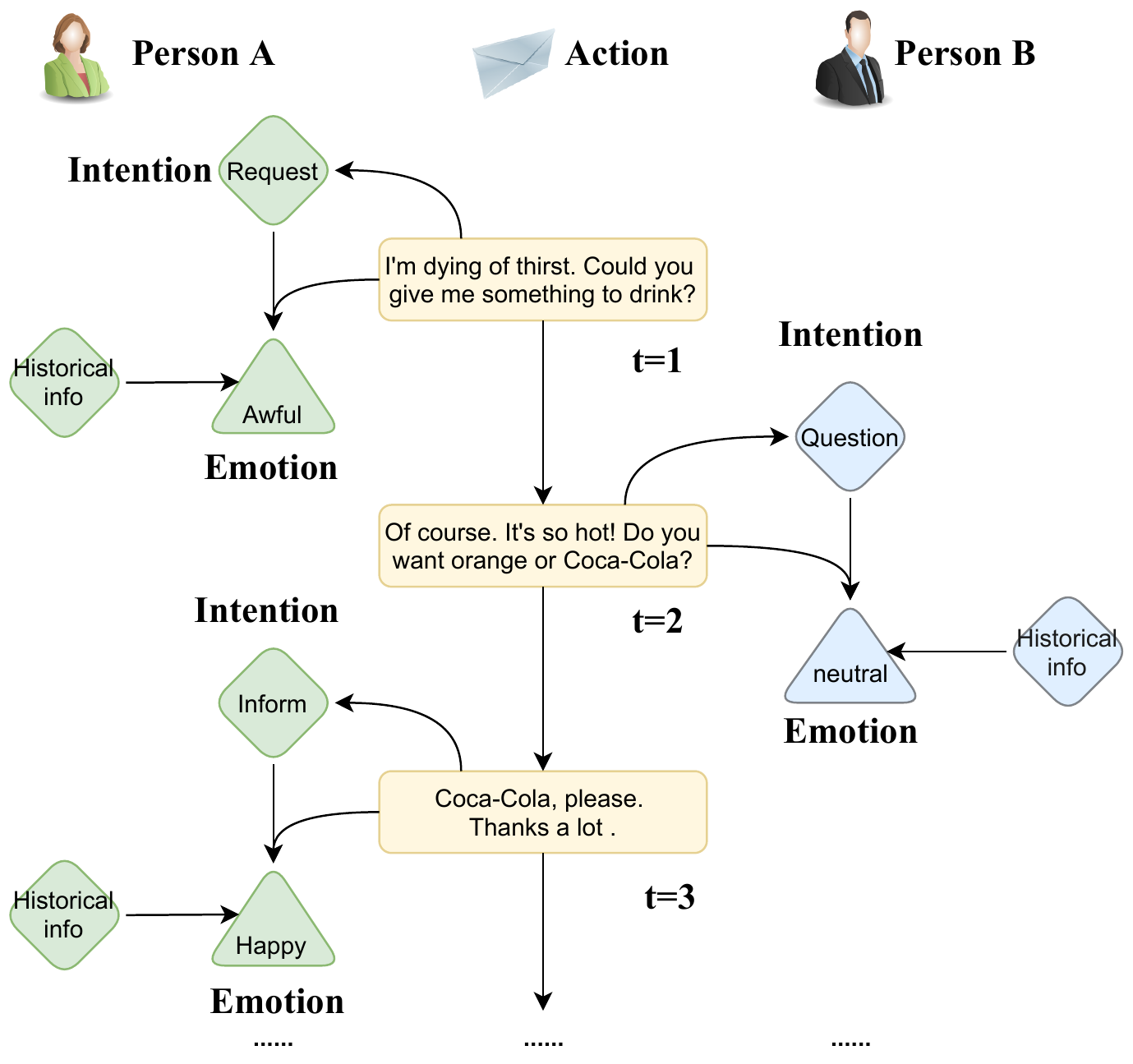}
	\caption{An example of dialogue systems. Diamond squares indicate intention, and different colors indicate different individuals. The triangle represents the emotion, and the rectangle represents the interaction action in the external world.}
	\label{fig:example}
\end{figure}

When the state of the external world changes (the customer bought satisfactory goods) and the change satisfies the intention, it is intuitive the customer is happy. As the work \cite{1989Emergent} demonstrated, emotion is the psychological behavior produced by the joint stimulation of the internal and external world. Namely, emotion is determined by both their own intention and external action. It often shows positive emotion when the action satisfies the intention \cite{1989Emergent,2006The}. Figure \ref{fig:example} shows the relationships between the intention, emotion and the action between the speaker and listener. Person A is thirst and wants to drink, so he puts forward a request to satisfy his intention \textit{request for drinking}. From the current intention and utterance, his emotion can be inferred as awful. Then, a new action \textit{Do you want orange or Coca-Cola} is triggered by person B from which the intention is inferred to \textit{question}. Similarly, person A expresses a new intention \textit{inform} by the utterance \textit{Coca-Cola, please}. And the emotion has been changed to \textit{happy} because of the satisfaction of historical intention \textit{request for drinking}. Thus, it’s critical to take the mutual relationships between the intention and emotion into account in an explicit way.
% During the above process, person A and person B interact with each other by actions \cite{2004A} and express their intentions and emotions by an implicit way. 

% a {1937 treaty} prohibiting the hunting of right and gray whales, and the {Bald Eagle Protection Act of 1940}.
\begin{figure*}[t]
	\centering
	\includegraphics[width=0.82\textwidth]{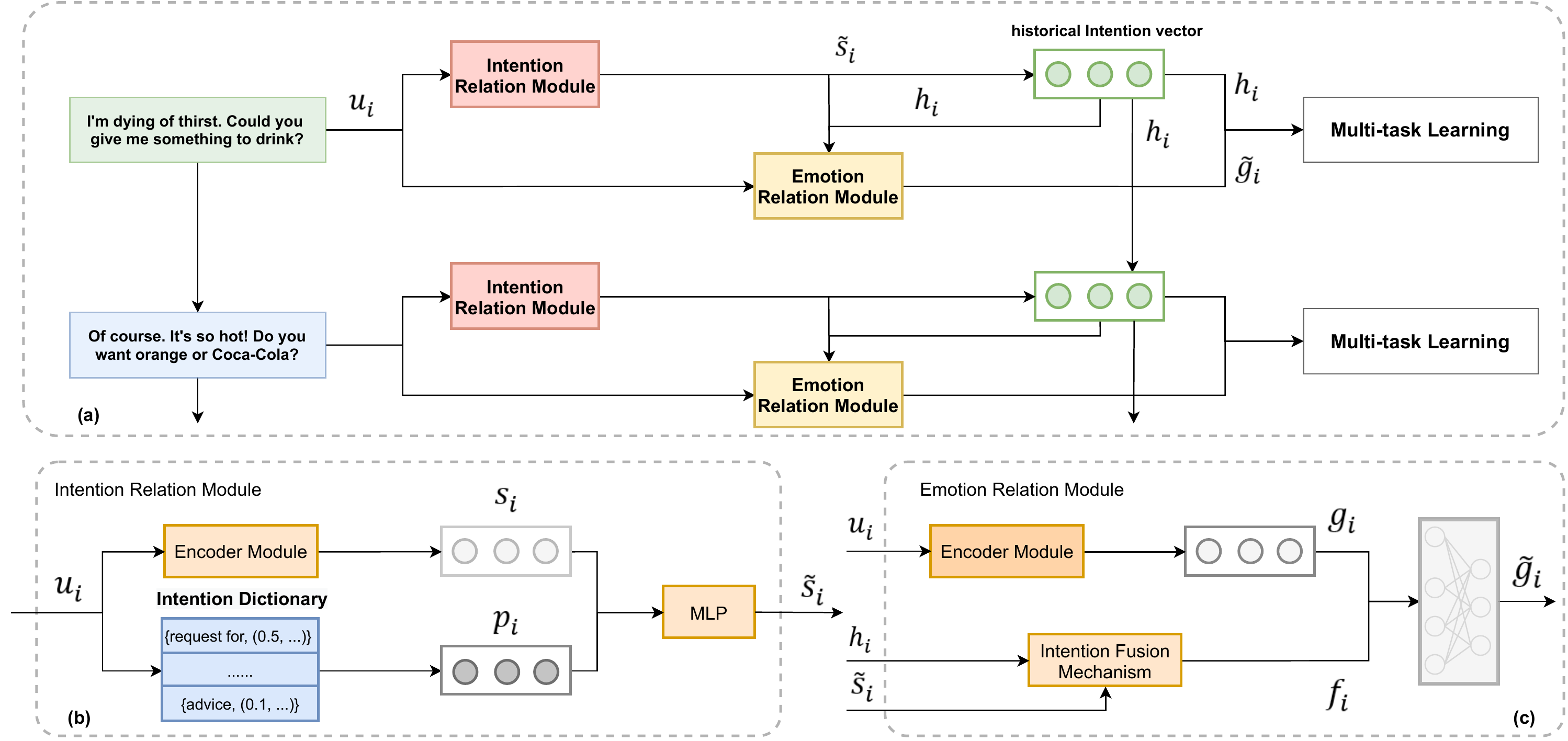} % Reduce the figure size so that it is slightly narrower than the column. Don't use precise values for figure width.This setup will avoid overfull boxes. 
	\caption{(a) is the overview of our RAIN which consists of the Intention Relation Module (b) and Emotion Relation Module (c).}
	\label{fig0}
\end{figure*}
Previous work \cite{DBLP:conf/emnlp/KumarAJ19,DBLP:conf/aaai/MajumderPHMGC19,DBLP:conf/aaai/ColomboCMVVC20} mainly focused on intention recognition or emotion prediction, but seldom considered these factors together and ignored to explicitly integrate historical intention information. Although some work \cite {DBLP:conf/coling/CerisaraJOL18,DBLP:journals/prl/KimK18,DBLP:conf/aaai/QinCLN020} proposed modeling the interaction between the intention recognition task and emotion prediction task, they just predicted the label by the representation of utterances or self attention mechanism and regarded the two tasks as the sentence classification task. These works also lack analysis and explanation of intention recognition and emotion prediction.

In this paper, we propose a \textbf{RelAtion Interaction Network (RAIN)}, consisting of {Intention Relation Module} and {Emotion Relation Module}, to jointly model mutual relationships and explicitly integrate historical intention information.
% to jointly model mutual relationships between intention and emotion, and explicitly integrate historical intention information to the Emotion Relation Module. 
Specifically, \textbf{Intention Relation Module} introduces an intention dictionary to explicitly account for the intention recognition task. Then, \textbf{Emotion Relation Module} designs an intention fusion mechanism to explicitly integrate historical intention information for subsequent emotion prediction.
% The contributions can be summarized as follows. We proposed the MCR-Net to explicitly model the mutual interaction and locate the key clues from coarse to fine in a Multi-step reasoning fashion. And the novel 
The important observation is the significant performance obtained from RAIN; it not only achieves the high performance on the dataset but also makes an explanation of the two tasks.

\section{METHODOLOGY}
\label{sec:METHODOLOGY}

As shown in Figure \ref{fig0} (a), the proposed model consists of the {Intention Relation Module} and {Emotion Relation Module}. First, the \textbf{Intention Relation Module} (b) obtains the contextual representations of dialogues with the econder module and introduces an intention dictionary to make an interpretable intention recognition. Then, the \textbf{Emotion Relation Module} (c) designs an intention fusion mechanism to explicitly integrate historical intention information. Finally, RAIN makes subsequent predictions with the intention vectors $\boldsymbol{\tilde{s}_i}$ and emotion vectors $\boldsymbol{\tilde{g}_i}$ by multi-task learning.
\subsection{Intention Relation Module}
\label{sec:plm}
\noindent
\textbf{Encoder Module.}
Considering the strong performance of pre-trained language models (PLMs) \cite{Vaswani2017AttentionIA,DBLP:journals/corr/abs-1909-11942,DBLP:journals/corr/abs-1907-11692} such as RoBERTa, we use them as the Encoder Module to obtain the contextual representations. Given a conversation $C = (u_1, u_2, \dots, u_N)$ a set of N utterances and $u_i = (x_{i,1}, x_{i,2}, \dots, x_{i,T})$ that consists of a sequence of T words, with $Y^s = (y_1^s, y_2^s, \dots, y_N^s)$ and $Y^e = (y_1^e, y_2^e, \dots, y_N^e)$ being the corresponding intention and emotion labels. To obtain the $i^{th}$ sentence representation, we consider using the hidden state corresponding to [CLS], as:
\begin{equation}\label{eq:0}
\boldsymbol{s_i} \!=\! {\rm RoBERTa}~({[CLS]}, x_{i,1}, \dots, x_{i,T}, {[SEP]})[0]
\end{equation}
\noindent
\textbf{Intention Dictionary.}
Observing that certain specific words such as \textit{ask for, proposal} can reflect intention, we extract the feature words in the process of labeling the feature. For example, \textit{request for, would like} has a high probability of being \textit{request}.
% and \textit{proposal, recommendation, advice} has a high probability of being \textit{suggest}. 
We count the number of labeled feature words, count the word frequency of each feature word over different intentions, and normalize it to get the probability distribution $\boldsymbol{p_i}$. Therefore, we construct an Intention Dictionary as the prior intention knowledge base (Figure \ref{fig0} (b)) and then output a probability distribution $\boldsymbol{p_i}$ of the keyword over all the corresponding intentions, which can be regarded as an interpretable signal to enhance the semantic of the intention.

Finally, the intention vectors $\boldsymbol{ \tilde{s}_i}$, integrating symbolic representation and neural representation, can be obtained as:
\begin{equation}\label{eq:1}
\boldsymbol{\tilde{s}_i} \!=\! {\rm MLP}~({\rm ReLU}(W_s^T \boldsymbol{s_i} + \boldsymbol{p_i}))
\end{equation}
where $\boldsymbol{ \tilde{s}_i}\in\mathbb{R}^{h}$, $W_s\in\mathbb{R}^{h \times l_s}$, $h$ is the dimension of the hidden state, $l_s$ is the number of the intention labels.

\subsection{Emotion Relation Module}
In this module, the inputs includes: the current utterance $u_i$, the historical intention information $\boldsymbol{h_i}$ and the output of the Intention Relation Module $\boldsymbol{\tilde{s}_i}$. The details are as follows.
% The emotion vectors $\boldsymbol{\tilde{g}_i}$ which integrate historical intention information are used for 

\noindent
\textbf{Encoder Module.} As shown in Figure \ref{fig0} (c), the Emotion Relation Module utilizes another RoBERTa \cite{DBLP:journals/corr/abs-1907-11692} as the Encoder Module for obtaining the sentence-level representation $\boldsymbol{g_i}$ with the same formulation in Equation (\ref{eq:0}), where $\boldsymbol{g_i}\in\mathbb{R}^{h}$.

\noindent
\textbf{Historical Intention Modeling.} As demonstrated in Section \ref{sec:intro}, emotion is the psychological behavior produced by the joint stimulation of the intention and action. To model mutual relationships between the intention and emotion, the proposed model introduces historical intention information $\boldsymbol{h_i}$. Considering the sequence modeling, we utilize the LSTM \cite{Hochreiter1997LongSM} to capture the temporal features within the intentions, as:
\begin{equation}\label{eq:2}
\boldsymbol{h_i} \!=\! {\rm LSTM}~(\boldsymbol{\tilde{s}_i}, \boldsymbol{h_{i-1}}))
\end{equation}

\begin{table}[!]	
	\centering
	\setlength\tabcolsep{10pt}%调列距
	%\begin{tabular*}{\hsize}{@{}@{\extracolsep{\fill}}l|cc|cc@{}}
	\resizebox{0.99\linewidth}{!}{
		\begin{tabular}{l|ccc|ccc}
			\toprule
			\multicolumn{1}{c|}{\multirow{2}{*}} & \multicolumn{3}{c|}{{Intention Recognition}}       & \multicolumn{3}{c}{{Emotion Prediction}}      \\  % \cline{2-7} 
			\multicolumn{1}{c|}{} 	& {P $\uparrow$}	& {R $\uparrow$}	& {F1 $\uparrow$}  	& {P $\uparrow$} & {R $\uparrow$}     & {F1 $\uparrow$}    \\ \midrule
			{GRU} \cite{DBLP:journals/corr/ChungGCB14}	 	& {46.18}	& {41.31}  & {43.61} 	& {42.25} & {41.64}  & {41.94}   \\
			{GRU+Attention} \cite{Vaswani2017AttentionIA}		& 48.66	& 41.43	& 44.75 	& 43.74   & 41.20   & 42.43    \\
			% - Main Point	& 77.63	& -0.42	& 50.40	& -0.16                \\
			{DCR-Net} $\dagger$ \cite{DBLP:conf/aaai/QinCLN020}		& 52.35	& 48.56	& 50.38  	& 47.24		& 43.91	& 45.51	\\ 
			{BERT} \cite{DBLP:conf/naacl/DevlinCLT19}		& 65.84	& 65.18	& 65.51 	& 55.35		& 55.42	& 55.38	\\ 
			\midrule
			%- APC \& CL-Task	& 77.12   & -0.83   & 49.64  	& -0.92  \\	
			{RoBERTa$_{base}$} \cite{DBLP:journals/corr/abs-1907-11692}  & 68.14   & 68.53 	& 68.33 	& 56.82 	& 57.89	& 57.35	 \\	
			{RoBERTa$_{large}$} \cite{DBLP:journals/corr/abs-1907-11692}		& {71.36}	& {70.47}	& {70.91}	& {59.51} & {58.77} & {59.13}  \\
			\textbf{RAIN} 	& \textbf{73.22}	& \textbf{72.64}	& \textbf{72.93}	& \textbf{65.35} & \textbf{62.84} & \textbf{64.07}  \\ \bottomrule	
	\end{tabular}}	
	\caption{\label{tab:main} Experimental results on the testset for tasks of intention recognition and emotion prediction. $\dagger$ indicates that the performance is reimplemented by ourselves.}
\end{table}

\noindent
\textbf{Intention Fusion Mechanism.}
The fusion mechanism \cite{DBLP:conf/coling/PengHXXYSW20,DBLP:conf/acl/MouMLX0YJ16} is a general approach that is model-independent, which focus on the mutual relationships between the two sources. Motivated by the work \cite{DBLP:conf/icassp/Peng0YXXZS21}, the intention fusion mechanism is proposed to effectively integrate historical intention information. Specifically, the fusion layer first utilizes a particular fusion unit to combine the representations between the intention vectors $\boldsymbol{\tilde{s}_i}$ and the historical intention information $\boldsymbol{h_i}$. 

\begin{equation} \label{equ:9} 
\boldsymbol{f_i}\!=\! {\rm Fuse}~(\boldsymbol{\tilde{s}_i}, \boldsymbol{h_i}) 
\end{equation}
where $\rm Fuse(\cdot, \cdot)$ is a typical fusion kernel.

The simplest fusion kernel can be a concatenation or addition operation of the two sources, followed by a linear or non-linear transformation. Generally, a heuristic matching trick with difference and element-wise product is found effective in combining different representations \cite{DBLP:conf/coling/PengHXXYSW20,DBLP:conf/acl/MouMLX0YJ16}:
\begin{equation} \label{equ:30} 
{\rm {Fuse}}(\boldsymbol{\tilde{s}_i}, \boldsymbol{h_i})={\rm {tanh}}(W_{f}^T[\boldsymbol{\tilde{s}_i};\boldsymbol{h_i};\boldsymbol{\tilde{s}_i} \circ \boldsymbol{h_i}; \boldsymbol{\tilde{s}_i}-\boldsymbol{h_i}]+b_{f})
\end{equation}
where $ W_{f} \in\mathbb{R}^{4h \times h}$, $ b_{f}\in\mathbb{R}^{h}$ are trainable parameters, $\circ$ denotes the element-wise product. The output dimension is projected back to the same size as the original representation $\boldsymbol{\tilde{s}_i}$ or $\boldsymbol{h_i}$. [;] indicates vector concatenation.

Finally, the emotion vectors $\boldsymbol{ \tilde{g}_i}$, modeling the mutual relationships between the intention and emotion, obtained as:
\begin{equation}\label{eq:g}
\boldsymbol{\tilde{g}_i} \!=\! {\rm ReLU}~( W_g^T [\boldsymbol{f_i}; \boldsymbol{g_i}] + b_{g} )
\end{equation}
where $\boldsymbol{ \tilde{g}_i}\in\mathbb{R}^{h}$, $W_g\in\mathbb{R}^{2h \times h}$ and $b_{g} \in\mathbb{R}^{h}$.

\subsection{Prediction for Intention and Emotion}
Intention Relation Module integrates symbolic representation and neural representation to output intention vectors $\boldsymbol{\tilde{s}_i}$. Emotion Relation Module models the mutual relationships between the intention and emotion, and fuses the historical intention information to output emotion vectors $\boldsymbol{\tilde{g}_i}$.
After obtaining the $\boldsymbol{\tilde{s}_i}$ and $\boldsymbol{\tilde{g}_i}$, we then consider two MLPs for performing intention recognition and emotion prediction, which is defined as:

\begin{equation}
\boldsymbol{ \hat{y}}_{i}^{m} = {\rm Softmax} ({W}_{m}^T{ \boldsymbol{\tilde{s}_i} } + {b}_{m}),
\end{equation}
\begin{equation}
\boldsymbol{ \hat{y}}_{i}^{e} = {\rm Softmax} ({W}_{e}^T{\boldsymbol{\tilde{g}_i} } + {b}_{e}),
\end{equation}
where $\boldsymbol{ \hat{y}}_{i}^{m}$, $\boldsymbol{ \hat{y}}_{i}^{e}$ are the predicted distribution for intention and emotion, ${W}_{m}^T \in\mathbb{R}^{h \times l_m}$, ${W}_{e}^T \in\mathbb{R}^{h \times l_e} $ are transformation matrices, $l_m$, $l_e$ are the number of intention and emotion labels.

%\begin{align} \label{equ:task2_loss}
%\mathcal{L}_{s}=-\frac{1}{N}\sum_{i=1}^{N}{\hat{y}} \log p_s	\\
%\mathcal{L}_{e}=-\frac{1}{N}\sum_{i=1}^{N}{y^s_e} \log p_e
%\end{align}

%\begin{table}[!]	
%	\centering
%	\setlength\tabcolsep{11pt}%调列距
%	%\begin{tabular*}{\hsize}{@{}@{\extracolsep{\fill}}l|cc|cc@{}}
%	\resizebox{0.98\linewidth}{!}{
%		\begin{tabular}{l|ccc|ccc}
%			\toprule
%			\multicolumn{1}{c|}{\multirow{2}{*}} & \multicolumn{3}{c|}{{Intention Recognition}}       & \multicolumn{3}{c}{{Emotion Prediction}}     \\  % \cline{2-7} 
%			\multicolumn{1}{c|}{} 	& {P $\uparrow$}	& {R $\uparrow$}	& {F1 $\uparrow$}  	& {P $\uparrow$} & {R $\uparrow$}     & {F1 $\uparrow$}     \\ \midrule
%			{RoBERTa$_{large}$}	 \cite{DBLP:journals/corr/abs-1907-11692}		& {71.36}	& {70.47}	& {70.91}	& {59.51} & {58.77} & {59.13}   \\	\midrule
%			~ \textbf{+IntentNet}		& {72.63}	& {71.95}	& {72.29}	& {61.02} & 61.28 & {61.15} \\
%			~\textbf{+Fusion Mechanism} & {-}	& {-}	& {-}	& {59.57} & {61.38} & {60.46}  \\
%			~\textbf{+Historical Intention}	& {72.10}	& {71.45}	& {71.77}	& {62.51} & {61.72} & {62.11}  \\
%			~\textbf{+Multi-task} 	& {72.26}	& {71.61}	& {71.96}	& {59.88} & {60.06} & {59.97}  \\	\midrule
%			~\textbf{RAIN} 	& \textbf{73.22}	& \textbf{72.64}	& \textbf{72.93}& \textbf{65.35} & \textbf{62.84} & \textbf{64.07} 	 \\	\bottomrule
%	\end{tabular}}
%	\caption{\label{tab:ablation} The results of ablation study on model components. }
%\end{table}
\begin{table}[!]	
	\centering
	\setlength\tabcolsep{11pt}%调列距
	%\begin{tabular*}{\hsize}{@{}@{\extracolsep{\fill}}l|cc|cc@{}}
	\resizebox{0.92\linewidth}{!}{
		\begin{tabular}{l|cc|cc}
			\toprule
			\multicolumn{1}{c|}{\multirow{2}{*}} & \multicolumn{2}{c|}{{Intention Recognition}}       & \multicolumn{2}{c}{{Emotion Prediction}}     \\  % \cline{2-7} 
			\multicolumn{1}{c|}{} 	& {F1 $\uparrow$} & $\Delta_{(F1)} $     & {F1 $\uparrow$}   & $\Delta_{(F1)} $  \\ \midrule
			{RoBERTa$_{large}$}	 \cite{DBLP:journals/corr/abs-1907-11692}		& {70.91}	& {-}  & {59.13} & {-}  \\	\midrule
			~ \textbf{+IntentNet}	& {72.29}	& {+1.38}  & {61.15} & {+2.02}\\
			~\textbf{+Fusion Mechanism} & {-}	& {-} & {60.46}  & {+1.33}\\
			~\textbf{+Historical Intention}	& {71.77}	& {+0.86}  & {62.11}  & {+2.98}\\
			~\textbf{+Multi-task} 	& {71.96}	& {+1.05}  & {59.97}  & {+0.84}\\	\midrule
			~\textbf{RAIN} 	& \textbf{72.93}& \textbf{+2.02} & \textbf{64.07} 	& \textbf{{+4.94}} \\	\bottomrule
	\end{tabular}}
	\caption{\label{tab:ablation} The results of ablation study on model components. }
\end{table}

The average cross-entropy loss of the intention recognition and emotion prediction are optimized as:
\begin{equation} \label{equ:l1}
\mathcal{L}_{m}=-\frac{1}{K}\sum_{j=1}^{K}\sum_{i=1}^{N}{{y_i^m}} \log \boldsymbol{ \hat{y}}_{i}^{m}
\end{equation}
\begin{equation} \label{equ:l2}
\mathcal{L}_{e}=-\frac{1}{K}\sum_{j=1}^{K}\sum_{i=1}^{N}{{y_i^e}} \log \boldsymbol{ \hat{y}}_{i}^{e}
\end{equation}
where $K$ is the total number of the examples, $N$ is the number of the utterances in one conversation, ${y_i^m}$ and ${y_i^e}$ are gold utterance intention label and gold emotion label.

% p(\mathbf{a} | \mathbf{H}^\text{r}) = p(a_\text{s} | \mathbf{H}^\text{r}) p(a_\text{e} | a_\text{s}, \mathbf{H}^\text{r}).\\

%\noindent
%\textbf{No-Answer Consistency Loss.} In \textbf{Answer Possibility Classifier}, if the score is larger than a certain threshold, the question can not be answered. At the same time, in \textbf{Answer Extraction}, if the probability of $0^{th}$ position it shows either. Therefore, we introduce a no-answer consistency loss rather than just depend on classifier which is more arbitrary.
%
%First, we have to construct another labels manually. If pointer point to $0^{th}$ position, label $\tilde{y}$ is ``1", else ``0". Similarity, the loss can be defined as:
%\begin{equation} \label{equ:12}
%% \mathcal{L}_{NACT}=-\frac{1}{N}\sum_{i=1}^{N}(y^{n_{i}} - score_i)^2
%\mathcal{L}_{NACT}=-\frac{1}{N}\sum_{i=1}^{N}[\tilde{y} \log \tilde{s}_{i} +(1-\tilde{y}) \log (1-\tilde{s}_{i})]
%\end{equation}
\noindent
\textbf{Joint Learning.} We combine the above two loss functions as the training loss in a multi-task learning manner 
% that jointly training can provide benefits and boost each other 
as:
%\begin{equation} \label{equ:13}
%\mathcal{L(\theta)}=\mathcal{L}_{AE}+\lambda_1\mathcal{L}_{APC}+\lambda_2\mathcal{L}_{NACT}
%\end{equation}
\begin{equation} \label{equ:13}
\mathcal{L(\theta)}=\lambda_1\mathcal{L}_{m}+\lambda_2\mathcal{L}_{e}
\end{equation}
where $\theta$ is the all learnable parameters, and $\lambda_1$ and $\lambda_2$ are two hyper-parameters for controlling the weight of the rest tasks.

% Because of the special encoding method of BERT, we make a limitation that answer span only selects from the passage to avoid extracting answer from the question. And the end position should be larger or equal to the start position.

\begin{figure*}[!]
	\centering
	\includegraphics[width=0.64\textwidth]{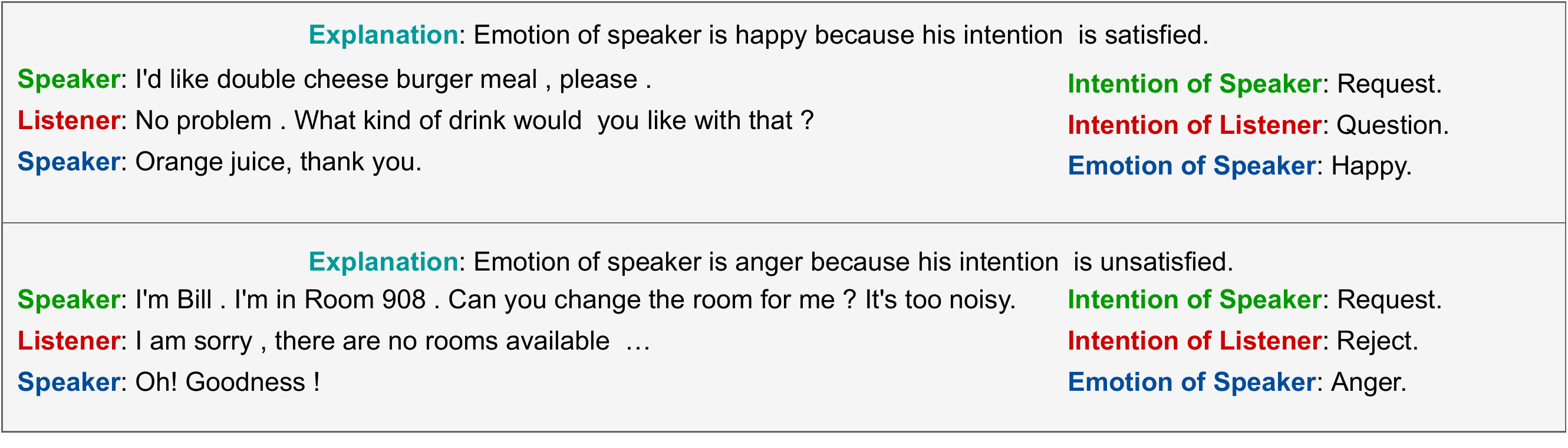}
	\caption{Two examples RAIN generated, we also give the explanation of the emotion prediction task.}
	\label{fig:case}
\end{figure*}
\begin{figure*}[!]
	\centering
	\includegraphics[width=0.88\textwidth]{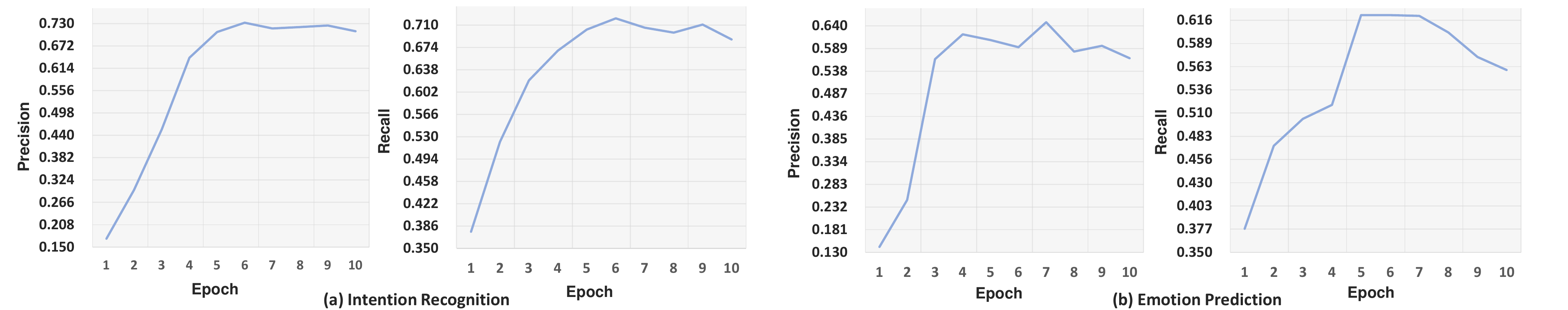}
	\caption{The performance with respect to the number of epochs on the dataset.}
	\label{fig:epoch}
\end{figure*}
\section{Experiments}
\label{sec:pagestyle}
\subsection{Experimental Setting}
\noindent
\textbf{Datasets \& Evaluation Metrics.} The DailyDialog \cite{DBLP:conf/ijcnlp/LiSSLCN17} datasets contain 13,118 multi-turn dialogues. In order to obtain the intention dictionary, we extract 2,046 conversations for the annotation. Furthermore, the intention classification has been expanded into seven categories (four categories in the original dataset), including \textit{request, suggest, command, accept, reject, question} and \textit{inform}. The emotion classification including \textit{happy, neutral, sadness, anger, content} and \textit{disgust}. As for the evaluation metric, macro-average Precision (P), Recall (R)
and F1 are considered for the DailyDialog dataset.

\noindent
\textbf{Implementation details.} The BERT-style baselines have same hyper parameters as \cite{DBLP:journals/corr/abs-1907-11692}. Additionly, we perform a grid search over the hyper-parameter settings (with a learning rate from \{0.01, 0.4\} for GRU or \{1e-5, 3e-5\} for PLMs (in Sec. \ref{sec:plm}), a batch size from \{16, 32\}, and epochs from \{3, 12\}). The hyper-parameters in the loss are $\lambda_1$ = $\lambda_2$ = 0.5. Models are trained to minimize the cross entropy with Adam \cite{DBLP:journals/corr/KingmaB14}.

\subsection{Experimental Results}
\noindent
\textbf{State-of-the-art Comparison.} We provide some baselines and proposed model for the two tasks. For the intention recognition, as shown in the first column of Table \ref{tab:main}, in all models, RoBERTa performs the best. Compared with the baselines, RAIN can outperform them by a large margin which achieves 2.02\% gain on F1. For the emotion prediction, as shown in the second column of Table \ref{tab:main}, RAIN achieves the best results that outperform the previous best-published model.
% which is a significantly improvement on F1 score.
% But all in all, the MCR-Net outperforms the previous methods.
% In DuReader, we also get high performance on the base model and large model.

\noindent
\textbf{Ablation Study.} To get a better insight into our proposed model, we perform the ablation study as shown in Table \ref{tab:ablation}. We add the proposed modules to explore their contribution. For comparison, the RoBERTa$ _{large}$ baseline and ours that utilizes all the components are also provided.

From the Table \ref{tab:ablation}, we can see that:
% We mainly focus on the EM and ROUGE-L score on two datasets. 
1) After utilizing the intention dictionary, the performance has improved on the two tasks. It shows that the prior statistical knowledge is effective to predict the intention as well as the emotion. 2) The fusion mechanism also has an improvement, proving that such an operation is indeed effective. 3) The historical intention information is beneficial for the two tasks, which validate the necessity of the historical modeling. 4) The multi-task learning can provide benefits, which show that the two training objectives are actually closely related and can boost each other. The ablation study has demonstrated that all the components proposed are beneficial for the tasks.

\noindent
\textbf{Case Study.}
In this section, we present dialogue cases and explanations of the emotion to demonstrate how our model performs. As shown in Fig. \ref{fig:case}, for the first case, the speaker requests for \textit{burger meal} and intention of the listener is \textit{question}. Our model produces a reasonable intention and emotion prediction. As for the explanation of the emotion, we give the generative template based on the outputs, like \textit{Emotion of speaker is happy because his intention is satisfied}.

\noindent
\textbf{Performance on the Number of Epochs.} To explore the influence of different epochs, another experiment with the proposed model is conducted. The comparison of the results is depicted in Fig. \ref{fig:epoch}. From the experiments, the conclusion is that during the first \{5, 6\} epochs, performances achieve a lot and reach a peak, while after that, performances drop slightly. 
% The models gain the best when epoch is set to 6 on the proposed two tasks. 
It demonstrates that the models can converge quickly during the first few epochs and capture the features between the utterances effectively.

\section{CONCLUSION}
\label{sec:foot}

In this paper, we present a RelAtion Interaction Network (RAIN) to jointly model mutual relationships between the intention and emotion, and explicitly integrate historical intention information. We show that the proposed RAIN is effective and interpretable, which outperforms the previous methods with a single model, as well as making an explanation of the two tasks. For the future work, some other psychological factors will be considered, such as personal character, educational background and so on. We believe that these cognitive factors are still worth researching for human activities.
% The ablation analyses demonstrate the importance of each component in our model, especially the Evidence Refining Reasoner.
%\section{Acknowledgments}
%This work is supported by the National Natural Science Foundation of China (No.62006222). 

%\vfill\pagebreak

% References should be produced using the bibtex program from suitable
% BiBTeX files (here: strings, refs, manuals). The IEEEbib.bst bibliography
% style file from IEEE produces unsorted bibliography list.
% -------------------------------------------------------------------------
\bibliographystyle{IEEEbib}
\bibliography{strings,refs}

\begin{thebibliography}{10}

\bibitem{1997Affective}
Rosalind~W. Picard,
\newblock {\em Affective computing},
\newblock MIT Press, 1997.

\bibitem{maslow2013theory}
Abraham~Harold Maslow,
\newblock {\em A theory of human motivation},
\newblock Simon and Schuster, 2013.

\bibitem{Gable2010Approach}
Philip~A. Gable and Eddie Harmon-Jones,
\newblock ``Approach-motivated positive affect reduces breadth of attention,''
\newblock {\em Psychological Science}, vol. 19, no. 5, pp. 476--482, 2010.

\bibitem{reeve2014understanding}
Johnmarshall Reeve,
\newblock {\em Understanding motivation and emotion},
\newblock John Wiley \& Sons, 2014.

\bibitem{1989Emergent}
Joseph~J. Campos, Rosemary~G. Campos, and Karen~C. Barrett,
\newblock ``Emergent themes in the study of emotional development and emotion
  regulation.,''
\newblock {\em Developmental Psychology}, vol. 25, no. 3, pp. 394--402, 1989.

\bibitem{2006The}
E.~J. Lawrence, P.~Shaw, V.~P. Giampietro, S.~Surguladze, M.~J. Brammer, and
  A.~S. David,
\newblock ``The role of 'shared representations' in social perception and
  empathy: an fmri study.,''
\newblock {\em Neuroimage}, vol. 29, no. 4, pp. 1173--1184, 2006.

\bibitem{DBLP:conf/emnlp/KumarAJ19}
Harshit Kumar, Arvind Agarwal, and Sachindra Joshi,
\newblock ``A practical dialogue-act-driven conversation model for multi-turn
  response selection,''
\newblock in {\em EMNLP-IJCNLP}, 2019, pp. 1980--1989.

\bibitem{DBLP:conf/aaai/MajumderPHMGC19}
Navonil Majumder, Soujanya Poria, Devamanyu Hazarika, Rada Mihalcea,
  Alexander~F. Gelbukh, and Erik Cambria,
\newblock ``Dialoguernn: An attentive {RNN} for emotion detection in
  conversations,''
\newblock in {\em AAAI}, 2019, pp. 6818--6825.

\bibitem{DBLP:conf/aaai/ColomboCMVVC20}
Pierre Colombo, Emile Chapuis, Matteo Manica, Emmanuel Vignon, Giovanna Varni,
  and Chlo{\'{e}} Clavel,
\newblock ``Guiding attention in sequence-to-sequence models for dialogue act
  prediction,''
\newblock in {\em AAAI}, 2020, pp. 7594--7601.

\bibitem{DBLP:conf/coling/CerisaraJOL18}
Christophe Cerisara, Somayeh Jafaritazehjani, Adedayo Oluokun, and Hoa~T. Le,
\newblock ``Multi-task dialog act and sentiment recognition on mastodon,''
\newblock in {\em COLING}, 2018, pp. 745--754.

\bibitem{DBLP:journals/prl/KimK18}
Minkyoung Kim and Harksoo Kim,
\newblock ``Integrated neural network model for identifying speech acts,
  predicators, and sentiments of dialogue utterances,''
\newblock {\em Pattern Recognit. Lett.}, vol. 101, pp. 1--5, 2018.

\bibitem{DBLP:conf/aaai/QinCLN020}
Libo Qin, Wanxiang Che, Yangming Li, Minheng Ni, and Ting Liu,
\newblock ``Dcr-net: {A} deep co-interactive relation network for joint dialog
  act recognition and sentiment classification,''
\newblock in {\em AAAI}, 2020, pp. 8665--8672.

\bibitem{Vaswani2017AttentionIA}
Ashish Vaswani, Noam Shazeer, Niki Parmar, Jakob Uszkoreit, Llion Jones,
  Aidan~N. Gomez, L.~Kaiser, and Illia Polosukhin,
\newblock ``Attention is all you need,''
\newblock {\em ArXiv}, vol. abs/1706.03762, 2017.

\bibitem{DBLP:journals/corr/abs-1909-11942}
Zhenzhong Lan, Mingda Chen, Sebastian Goodman, Kevin Gimpel, Piyush Sharma, and
  Radu Soricut,
\newblock ``{ALBERT:} {A} lite {BERT} for self-supervised learning of language
  representations,''
\newblock {\em CoRR}, vol. abs/1909.11942, 2019.

\bibitem{DBLP:journals/corr/abs-1907-11692}
Yinhan Liu, Myle Ott, Naman Goyal, Jingfei Du, Mandar Joshi, Danqi Chen, Omer
  Levy, Mike Lewis, Luke Zettlemoyer, and Veselin Stoyanov,
\newblock ``Roberta: {A} robustly optimized {BERT} pretraining approach,''
\newblock {\em CoRR}, vol. abs/1907.11692, 2019.

\bibitem{Hochreiter1997LongSM}
S.~Hochreiter and J.~Schmidhuber,
\newblock ``Long short-term memory,''
\newblock {\em Neural Computation}, vol. 9, pp. 1735--1780, 1997.

\bibitem{DBLP:journals/corr/ChungGCB14}
Junyoung Chung, {\c{C}}aglar G{\"{u}}l{\c{c}}ehre, KyungHyun Cho, and Yoshua
  Bengio,
\newblock ``Empirical evaluation of gated recurrent neural networks on sequence
  modeling,''
\newblock {\em CoRR}, vol. abs/1412.3555, 2014.

\bibitem{DBLP:conf/naacl/DevlinCLT19}
Jacob Devlin, Ming{-}Wei Chang, Kenton Lee, and Kristina Toutanova,
\newblock ``{BERT:} pre-training of deep bidirectional transformers for
  language understanding,''
\newblock in {\em NAACL-HLT}, 2019, pp. 4171--4186.

\bibitem{DBLP:conf/coling/PengHXXYSW20}
Wei Peng, Yue Hu, Luxi Xing, Yuqiang Xie, Jing Yu, Yajing Sun, and Xiangpeng
  Wei,
\newblock ``Bi-directional cognitivethinking network for machine reading
  comprehension,''
\newblock in {\em COLING}, Donia Scott, N{\'{u}}ria Bel, and Chengqing Zong,
  Eds. 2020, pp. 2613--2623, International Committee on Computational
  Linguistics.

\bibitem{DBLP:conf/acl/MouMLX0YJ16}
Lili Mou, Rui Men, Ge~Li, Yan Xu, Lu~Zhang, Rui Yan, and Zhi Jin,
\newblock ``Natural language inference by tree-based convolution and heuristic
  matching,''
\newblock in {\em ACL}. 2016, The Association for Computer Linguistics.

\bibitem{DBLP:conf/icassp/Peng0YXXZS21}
Wei Peng, Yue Hu, Jing Yu, Luxi Xing, Yuqiang Xie, Zihao Zhu, and Yajing Sun,
\newblock ``{MCR-NET:} {A} multi-step co-interactive relation network for
  unanswerable questions on machine reading comprehension,''
\newblock in {\em ICASS}. 2021, pp. 7818--7822, {IEEE}.

\bibitem{DBLP:conf/ijcnlp/LiSSLCN17}
Yanran Li, Hui Su, Xiaoyu Shen, Wenjie Li, Ziqiang Cao, and Shuzi Niu,
\newblock ``Dailydialog: {A} manually labelled multi-turn dialogue dataset,''
\newblock in {\em IJCNLP}, 2017, pp. 986--995.

\bibitem{DBLP:journals/corr/KingmaB14}
Diederik~P. Kingma and Jimmy Ba,
\newblock ``Adam: {A} method for stochastic optimization,''
\newblock in {\em ICLR}, 2015.

\end{thebibliography}

\end{document}